\documentclass[runningheads]{llncs}
\usepackage{graphicx}
\usepackage{bbding}
\usepackage{amsmath,amssymb}
\usepackage{mathrsfs}
\usepackage{bm}
\usepackage{booktabs}
\usepackage{multirow}
\usepackage[dvipsnames]{xcolor}
\usepackage[colorlinks,linkcolor=blue,anchorcolor=blue,citecolor=blue]{hyperref}

\begin{document}
\title{Swin Deformable Attention U-Net Transformer (SDAUT) for Explainable Fast MRI}

% \orcidID{***}
\author{Jiahao Huang \inst{1, 2}\and
Xiaodan Xing \inst{1} \and
Zhifan Gao \inst{3} \and
Guang Yang \inst{1, 2} \Envelope}
% index{Huang, Jiahao}
% index{Xing, Xiaodan} 
% index{Gao, Zhifan} 
% index{Yang, Guang} 

\authorrunning{Huang et al.}
\titlerunning{SDAUT for Explainable Fast MRI}

\institute{
National Heart and Lung Institute, Imperial College London, London, United Kingdom \\
\email{g.yang@imperial.ac.uk} \and
Cardiovascular Research Centre, Royal Brompton Hospital, London, United Kingdom \and
School of Biomedical Engineering, Sun Yat-sen University, Guangzhou, China}

\maketitle              % typeset the header of the contribution

\begin{abstract}
%The abstract should briefly summarize the contents of the paper in 15--250 words.
Fast MRI aims to reconstruct a high fidelity image from partially observed measurements. Exuberant development in fast MRI using deep learning has been witnessed recently. Meanwhile, novel deep learning paradigms, e.g., Transformer based models, are fast-growing in natural language processing and promptly developed for computer vision and medical image analysis due to their prominent performance. Nevertheless, due to the complexity of the Transformer, the application of fast MRI may not be straightforward. The main obstacle is the computational cost of the self-attention layer, which is the core part of the Transformer, can be expensive for high resolution MRI inputs. In this study, we propose a new Transformer architecture for solving fast MRI that coupled Shifted Windows Transformer with U-Net to reduce the network complexity. We incorporate deformable attention to construe the explainability of our reconstruction model. We empirically demonstrate that our method achieves consistently superior performance on the fast MRI task. Besides, compared to state-of-the-art Transformer models, our method has fewer network parameters while revealing explainability. The code is publicly available at https://github.com/ayanglab/SDAUT. 

\end{abstract}

\section{Introduction}

% MRI
\par{Although magnetic resonance imaging (MRI) is widely utilised in clinical practice, it has been restricted by the prolonged scanning time. Fast MRI has relied heavily on image reconstruction from undersampled k-space data, e.g., using parallel imaging, simultaneous multi-slice, and compressed sensing (CS)~\cite{Chen2022ai}. However, these conventional methods were suffered from limited acceleration factors or slow nonlinear optimisation. Deep learning has recently proven enormous success in a variety of research domains, as well as shown the capability to substantially accelerate MRI reconstruction with fewer measurements.}

\par{With its superior reconstruction quality and processing speed, convolutional neural networks (CNNs) based fast MRI methods~\cite{wang2016accelerating,Yang2018,yuan2020sara} enabled enhanced latent feature extraction by the deep hierarchical structure, and were successfully developed for a wide range of MRI sequences and clinical applications~\cite{schlemper2018stochastic,cheng2021learning,li2021high}.}

\par{Recently, taking advantage of sequence-to-sequence and deeper architectures, Transformers~\cite{Vaswani2017} showed superiority in computer vision tasks~\cite{Carion2020,Dosovitskiy2020,Parmar2018}, mainly ascribed to their larger receptive fields and long-range dependency~\cite{Parmar2018,Salimans2017} compared to their CNN counterparts. Transformer based methods were then swiftly developed in medical image analysis, e.g., for segmentation~\cite{Hatamizadeh_2022_WACV}, cross-modality synthesis~\cite{Shin2020}, and reconstruction~\cite{KorkmazDeep2021} with superior performance obtained.}

However, the dense attention design, e.g., in Vision Transformer (ViT)~\cite{Dosovitskiy2020}, was quite unfriendly to tasks working on high-resolution image inputs, e.g., MRI super-resolution and reconstruction, leading to excessive memory and computational costs. Moreover, visual elements tend to have a substantial variance in scale, and features could be influenced by the redundant attention~\cite{Liu2021}. Attempts using sparse attention, e.g., Shifted Windows (Swin) Transformer~\cite{Liu2021} significantly alleviated the memory and computational costs that could better fit the MRI reconstruction scenario~\cite{Huang2022_Data_and_Physics_Driven,Huang2022_SwinMR,Huang2022_STGAN}. Nevertheless, the selection of windows in Swin Transformer was data agnostic, which limited the capability of modelling long-range relationships and turned Swin Transformer into an \emph{attention-style CNN}. On the other hand, most existing deep learning based MRI reconstruction methods, including CNN and Transformer, are still black-boxes~\cite{yang2022unbox} that are intricate for interpretation by the end-users.

\begin{figure}[thpb]
    \centering
    \includegraphics[width=5in]{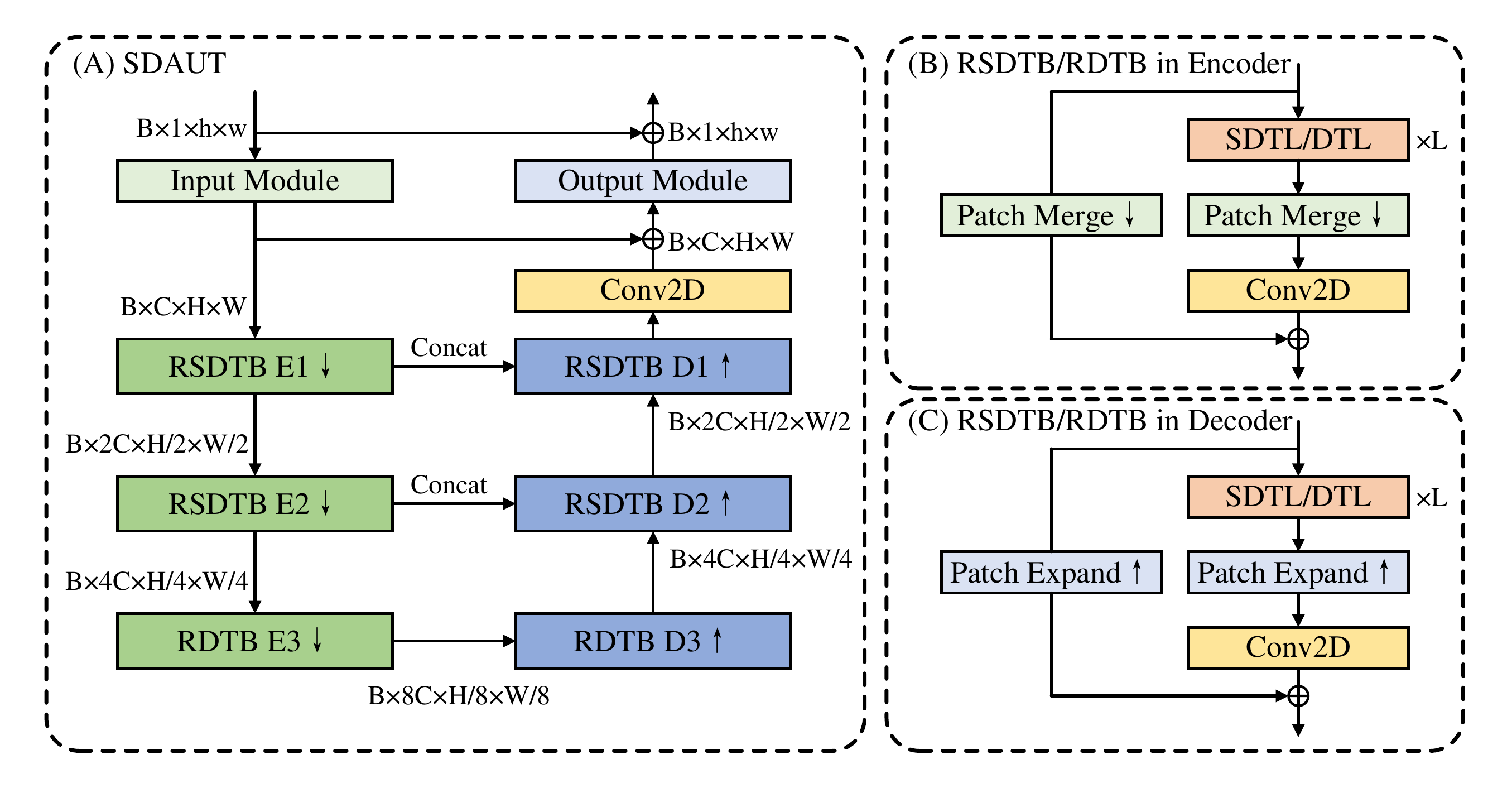}
    \caption{
    (A) Structure of our Swin Deformable Attention U-Net Transformer (SDAUT); (B) Structure of residual (Swin) deformable Transformer Block (RSDTB/RDTB) in the encoder path; (C) The structure of RSDTB/RDTB in the decoder path.
    }
    \label{fig:FIG_STRUCTURE}
\end{figure}

Inspired by the deformable convolution networks~\cite{Dai_2017_ICCV}, Xia et al.~\cite{Xia2022} proposed a ViT with deformable attention, namely DAT, in which the key and value were selected in a data-dependent way. DAT demonstrated advantages on classification and detection tasks; however, the dense attention design aggravated the memory and computational costs, making it difficult to be utilised in high-resolution MRI reconstruction.

In this work, we propose a Swin Deformable Attention U-Net Transformer (SDAUT) to combine dense and sparse deformable attention in different stages of a U-Net based Transformer. Sparse deformable attention is developed for shallow and high-resolution feature maps to preserve image details while reducing computational costs. Besides, dense deformable attention is leveraged for deep and low-resolution feature maps to exploit global information. This can also achieve long-range dependency at lower memory and computational costs. Furthermore, the utilisation of deformable attention can improve the reconstruction quality and enhance the explainability. We hypothesise that (1) the deformable attention Transformer can strengthen the performance of fast MRI with limited growth of computational cost; (2) the incorporation of U-Net architecture with both dense and sparse attention can reduce the number of network parameters significantly but improve the model performance; and (3) more importantly, our model is competent to achieve fast MRI while providing explainability.  

\section{Methods}

\subsection{U-Net Based Transformer}

Our proposed SDAUT adopted an end-to-end U-Net~\cite{Ronneberger2015} architecture, of which the input and output are undersampled zero-filled MRI images $x_u$ and reconstructed MR images $\hat x_u$, respectively. The SDAUT is composed of three modules (Fig.~\ref{fig:FIG_STRUCTURE}): the input module (IM), the U-Net based Transformer module (UTM), and the output module (OM), which can be expressed as $F_{\text{IM}} = \text{H}_{\text{IM}}(x_u)$, $F_{\text{UTM}} = \text{H}_{\text{UTM}}(F_{\text{IM}})$, and $F_{\text{OM}} = \text{H}_{\text{OM}}(F_{\text{UTM}}+F_{\text{IM}})$.

% IM & OM
\subsubsection{Input Module and Output Module}
The input module is a 2D convolutional layer (Conv2D) with the stride of the patch size $s$, which turns pixels in an area of $s \times s$ into a patch and maps the shallow inputs $x_u \in \mathbb{R}^{1 \times h \times w}$ to deep feature representations $F_{\text{IM}} \in \mathbb{R}^{C \times H \times W}$ ($H = h/s$ and $W = w/s$).

The output module is composed of a pixel shuffle Conv2D with the scale of the patch size $s$ and a Conv2D. This module recovers the pixels from patches and maps the deep feature representations $(F_{\text{IM}} + F_{\text{UTM}}) \in \mathbb{R}^{C \times H \times W}$ to shallow output images $F_{\text{OM}} \in \mathbb{R}^{1 \times h \times w}$. A residual connection from the input of the IM to the output of the OM turns the network into a refinement function $\hat x_u = F_{\text{OM}} + x_u$ to stable the training and accelerate the converge process. 

% UTM
\subsubsection{U-Net Based Transformer Module}
Our novel UTM comprises six cascaded residual Transformer blocks (Fig.~\ref{fig:FIG_STRUCTURE}), where two downsampling residual Swin deformable Transformer blocks (RSDTBs) and one downsampling residual deformable Transformer block (RDTB) are cascaded in the encoder path, and one upsampling RDTB and two upsampling RSDTBs are in the symmetric decoder path. Between the downsampling and upsampling Transformer blocks with the same scale, skip connection and concatenation are utilised to pass the information in the encoder path directly to the decoder path, for better feature preservation in different stages. 
Three cascaded downsampling Transformer blocks can be expressed as $F_{\text{E1}} = {\text{H}}_{\text{E1}}(F_{\text{IM}})$, $F_{\text{E1}} = {\text{H}}_{\text{E2}} (F_{\text{E1}})$ and $F_{\text{E3}} = {\text{H}}_{\text{E3}} (F_{\text{E2}})$.
Three cascaded upsampling Transformer blocks can be expressed as $F_{\text{D3}} = {\text{H}}_{\text{D3}} (F_{\text{E3}})$, $F_{\text{D2}} = {\text{H}}_{\text{D2}} (\operatorname{Concat}(F_{\text{D3}},F_{\text{E2}}))$ and $F_{\text{D1}} = {\text{H}}_{\text{D1}} (\operatorname{Concat}(F_{\text{D2}},F_{\text{E1}}))$.
The final Conv2D can be expressed as $F_{\text{UTM}} = {\text{H}}_{\text{CONV}}(F_{\text{D1}})$.

\subsubsection{Residual Swin Deformable Transformer Block}

As Fig.~\ref{fig:FIG_STRUCTURE} (B) and (C) shown, an RSDTB consists of $L$ cascaded Swin deformable Transformer layers (SDTLs) at the beginning and a Conv2D at the end. A residual connection is applied to connect the input and output of the RSDTB. 

For the downsampling RSDTBs in the encoder path, two patch merging layers~\cite{Liu2021} are placed behind the SDTLs and in the shortcut connection to reduce the patch number (patch resolution) and increase the channel number (depth).
For the upsampling RSDTBs in the decoder path, the patch merging layers are replaced by patch expending layers~\cite{Hu2021_SwinUnet} to enlarge the patch number (patch resolution) and reduce the channel number (depth). 
The RSDTB can be expressed as $F_{i,j} = {\text{H}}_{\text{SDTL}_{i,j}} (F_{i,j-1})\quad (j = 1,2,...,L)$, and $F_{i} = {\text{H}}_{\text{CONV}_{i}}({\text{H}}_{\text{S}_i}(F_{i,j})) + {\text{H}}_{\text{S}_i}(F_{i,0})$.
In the encoder, $i \in \{\text{E1}, \text{E2}, \text{E3}\}$ and ${\text{H}}_{\text{S}_i}$ denotes the patch merging layer. In the decoder, $i \in \{\text{D3}, \text{D2}, \text{D1}\}$ and ${\text{H}}_{\text{S}_i}$ denotes the patch expending layer.

\subsubsection{Swin Deformable Transformer Layer}

The Swin deformable Transformer layer adopts a standard Transformer layer structure, which can be expressed as $F_{i,j}^{\prime}={\text{H}}_{\text{SDMSA}}({\text{H}}_{\text{LN}}(F_{i,j-1}))+F_{i,j-1}$ and $F_{i,j}={\text{H}}_{\text{MLP}}({\text{H}}_{\text{LN}}(F_{i,j}^{\prime}))+F_{i,j}^{\prime}$. 
${\text{H}}_{\text{SDMSA}}$ denotes the Swin deformable multi-head self-attention, which is discussed in the next section. ${\text{H}}_{\text{LN}}$ and ${\text{H}}_{\text{MLP}}$ denote the Layer normalisation layer and multi-layer perceptron respectively.

\begin{figure}[ht]
    \centering
    \includegraphics[width=5in]{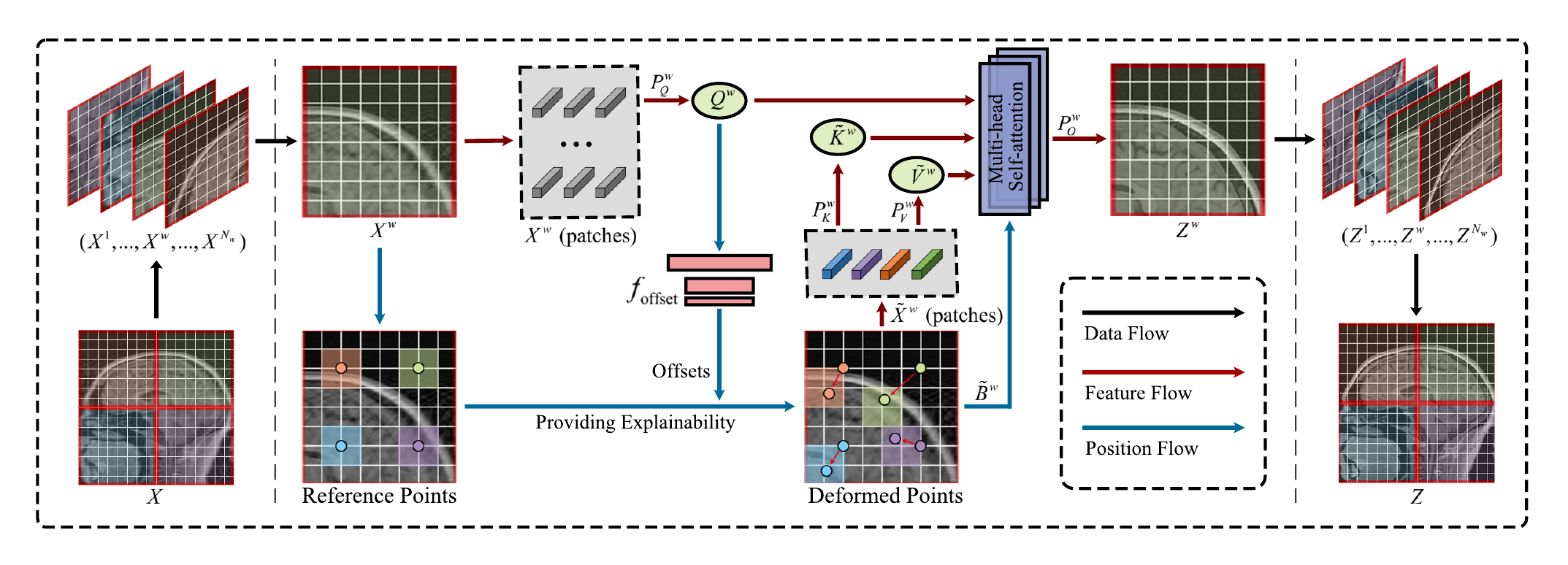}
    \caption{
    The structure of Swin deformable Transformer layer, which also provides model explainability. The subscript of the heads ($1,...,h,...,N_h$) is omitted for a clear illustration.
    }
    \label{fig:FIG_SDTL}
\end{figure}

\subsection{Shifted Windows Deformable Attention}

We propose a novel Swin deformable multi-head self-attention (SDMSA). Compared to deformable multi-head self-attention (DMSA) in DAT~\cite{Xia2022}, spatial constraints are applied in SDMSA (Fig.~\ref{fig:FIG_SDTL}). For an input feature map $X$, it is first split into $N_w$ windows spatially, and then divided into $N_h$ heads along the channels, which can be expressed as follows
\begin{align}\label{formula}
X = (X^{1},...,X^{w},...,X^{N_w}), \quad X^{w} = \operatorname{Concat}({X_{1}^{w},...X_{h}^{w},...,X_{N_h}^{w}}),
\end{align}

\noindent where $N_w=HW/W_s^2$ is the number of windows, and $N_h$ is the number of heads. For an input feature map $X_{h}^{w} \in \mathbb{R}^{C/N_h \times W_s \times W_s}$, a set of uniform grid points $p \in \mathbb{R}^{W_s/r \times W_s/r \times 2}$ are generated as the reference points with a downsampling factor $r$. A two-layer CNN $f_{\text{offset}}(\cdot)$ with GeLU layer is trained to learn the offsets $\Delta{p}$ of every reference point $p$. A bilinear interpolation function $\Phi(\cdot;\cdot)$ is utilised to sample the input feature map $X_{h}^{w}$ according to the deformed points $p+\Delta{p}$, and generate the sampled feature map $\widetilde{X}_{h}^{w} \in \mathbb{R}^{C/N_h \times W_s/r \times W_s/r}$. Details of $f_{\text{offset}}(\cdot)$ and $\Phi(\cdot;\cdot)$ can be found in~\cite{Xia2022}.

The deformable attention for the $w^{\text{th}}$ window $h^{\text{th}}$ head feature map $X_{h}^{w}$ can be expressed as follows
\begin{align}\label{formula:12}
& Q_{h}^{w}=X_{h}^{w} {P_{Q}}_{h}^{w}, \quad
\widetilde{K}_{h}^{w}=\widetilde{X}_{h}^{w} {P_{K}}_{h}^{w}, \quad
\widetilde{V}_{h}^{w}=\widetilde{X}_{h}^{w} {P_{V}}_{h}^{w}, \\
& \text{with} \quad 
\Delta{p} =a\operatorname{tanh}(f_{\text{offset}}(p)), \quad 
\widetilde{X}_{h}^{w} = \Phi({X}_{h}^{w};p+\Delta{p}), \\
& Z_{h}^{w} = \operatorname{Attention}(Q_{h}^{w}, \widetilde{K}_{h}^{w}, \widetilde{V}_{h}^{w})=\operatorname{SoftMax}\left(Q_{h}^{w} {\widetilde{K}_{h}^{w^T}} / \sqrt{d}+\widetilde{B}_{h}^{w}\right) \widetilde{V}_{h}^{w},
\end{align}

\noindent where ${P_{Q}}_{h}^{w}$ and ${P_{K}}_{h}^{w}$ and ${P_{V}}_{h}^{w}$ are the projection matrices (Conv2Ds were used in practice) for query $Q_{h}^{w}$, key $\widetilde{K}_{h}^{w}$, and value $\widetilde{V}_{h}^{w}$. Besides, $\widetilde{B}_{h}^{w}$ is the relative position bias, $d = C/N_h$ is the dimension of each head, and $a\operatorname{tanh}(\cdot)$ is used to avoid the overly-large offsets.

The attention outputs $Z_{h}^{w}$ of different heads are concatenated with a projection ${P_{O}}^{w}$. Then the output $Z^{w}$ of different windows are combined to generate the output $Z$ of SDMSA, which can be expressed as follows
\begin{align}\label{formula:13}
Z^{w} = \operatorname{Concat}({Z_{1}^{w},...Z_{h}^{w},...,Z_{N_h}^{w}}) {P_{O}}^{w}, \quad Z = (Z^{1},...,Z^{w},...,Z^{N_w}).
\end{align}

\subsection{Loss Function}

A content loss is applied to train our SDAUT, which is composed of a pixel-wise Charbonnier loss $\mathcal{L}_{\mathrm{pixel}}$, a frequency Charbonnier loss $\mathcal{L}_{\mathrm{freq}}$ and a perceptual VGG $l_1$ loss $\mathcal{L}_{\mathrm{VGG}}$ between reconstructed MR images $\hat x_u$ and ground truth MR images $x$ that totals $\mathcal{L}_{\mathrm{Total}}(\theta)
= \alpha \mathcal{L}_{\mathrm{pixel}}(\theta)
+ \beta \mathcal{L}_{\mathrm{freq}}(\theta)
+ \gamma \mathcal{L}_{\mathrm{VGG}}(\theta)$.

\section{Experimental Settings and Results}

\subsection{Implementation Details and Evaluation Methods}

We used the Calgary Campinas dataset~\cite{Souza2018} to train and test our methods. In total, 67 cases of 12-channel T1-weight 3D acquisitions were included in our study and randomly divided into training (40 cases), validation (7 cases) and independent testing (20 cases) in a ratio of 6:1:3 approximately. For each case, 100 2D sagittal-view slices near the centre were used. Root sum squared was applied to convert multi-channel data into single channel data.

Our proposed methods (except `KKDDKK-O-1' and `KKDDKK-NO-1') were trained for 100,000 steps with a batch size of 8, on two NVIDIA RTX 3090 GPUs with 24GB GPU RAM. The proposed methods with setting of `KKDDKK-O-1' or `KKDDKK-NO-1' were trained for 100,000 steps with a batch size of 2, on two NVIDIA Quadro RTX GPUs with 48GB GPU RAM. The setting naming rule is explained in Ablation Studies. For the network parameter, the layer number $L$, downsampling factor $r$ and window size $W_s$ were set to 6, 1 and 8. The channel number and head number of different blocks were set to $[90,180,360,720,360,180]$ and $[6,12,24,24,24,12]$.
The initial learning rate was set to $2 \times 10^{-4}$ and decayed every 10,000 steps by 0.5 from the 50,000$^{\text{th}}$ step. 
$\alpha$, $\beta$ and $\gamma$ in the total loss function were set to 15, 0.1 and 0.0025, respectively. 

Our proposed SDAUT was compared with other fast MRI methods, e.g., DAGAN~\cite{Yang2018}, nPIDD-GAN~\cite{Huang2021}, Swin-UNet~\cite{Hu2021_SwinUnet} and SwinMR~\cite{Huang2022_SwinMR} using Gaussian 1D 30\% and radial 10\% mask.

Peak Signal-to-Noise Ratio (PSNR), Structural Similarity Index Measure (SSIM), and Fr\'echet Inception Distance (FID)~\cite{Heusel2017} were used as evaluation metrics. Multiply Accumulate Operations (MACs) were applied to estimate the computational complexity for an input size of $1 \times 256 \times 256$.

\begin{table}[htbp]
  \centering
  \caption{Quantitative comparison results using a Gaussian 1D 30\% and radial 10\% mask. (SDAUT$_{x}$: SDAUT-KKDDKK-O-$x$; MACs of the generator without the discriminator in the GAN based model; $^{\dagger}$: $p < 0.05$ by paired t-Test compared with our proposed SDAUT$_{1}$; std: standard deviation.)}
    \scalebox{0.85}{
    \begin{tabular}{cccccccc}
        \toprule
        \multirow{2}[4]{*}{Method} & MACs  & \multicolumn{3}{c}{Gaussian 1D 30\%} & \multicolumn{3}{c}{Radial 10\%} \\
        \cmidrule{3-8}       & (G)   & SSIM $_\text{mean~(std)}$  & PSNR $_\text{mean~(std)}$  & FID   & SSIM $_\text{mean~(std)}$  & PSNR $_\text{mean~(std)}$  & FID \\
        \midrule
        ZF    & -     & 0.883 (0.012)$^{\dagger}$ & 27.81 (0.82)$^{\dagger}$ & 156.38 & 0.706 (0.022)$^{\dagger}$ & 23.53 (0.82)$^{\dagger}$ & 319.45 \\
        DAGAN & 33.97$^{*}$ & 0.924 (0.010)$^{\dagger}$ & 30.41 (0.82)$^{\dagger}$ & 56.04 & 0.822 (0.024)$^{\dagger}$ & 25.95 (0.85)$^{\dagger}$ & 132.58 \\
        nPIDD-GAN & 56.44$^{*}$ & 0.943 (0.009)$^{\dagger}$ & 31.81 (0.92)$^{\dagger}$ & 26.15 & 0.864 (0.023)$^{\dagger}$ & 27.17 (0.97)$^{\dagger}$ & 82.86 \\
        SwinUNet & 11.52 & 0.951 (0.008)$^{\dagger}$ & 32.52 (0.98)$^{\dagger}$ & 31.16 & 0.844 (0.027)$^{\dagger}$ & 26.41 (0.97)$^{\dagger}$ & 71.19 \\
        SwinMR & 800.73 & 0.955 (0.009)$^{\dagger}$ & 33.05 (1.09)$^{\dagger}$ & 21.03 & 0.876 (0.022)$^{\dagger}$ & 27.86 (1.02)$^{\dagger}$ & 59.01 \\
        \midrule
        SDAUT$_{2}$ & 57.91 & 0.956 (0.009)$^{\dagger}$ & 33.00 (1.12)$^{\dagger}$ & 22.54 & 0.867 (0.025)$^{\dagger}$ & 27.52 (1.05)$^{\dagger}$ & 62.92 \\
        SDAUT$_{1}$ & 293.02 & \textbf{0.963 (0.007)} & \textbf{33.92 (1.11)} & \textbf{20.45} & \textbf{0.885 (0.025)} & \textbf{28.28 (1.14)} & \textbf{55.12} \\
        \bottomrule
    \end{tabular}%
    }
  \label{tab:result}%
\end{table}%

\subsection{Comparison Studies}

Table~\ref{tab:result} shows the quantitative results of the comparison using a Gaussian 1D 30\% and radial 10\% mask. Results demonstrate that our proposed method outperforms other CNN based fast MRI methods, and achieves similar results compared to SwinMR but with significantly lower computation complexity.

\subsection{Ablation Studies}

We set four groups of ablation studies to investigate the influence of different block settings and utilisation of the deformable attention. We used `K' and `D' to denote the RSDTB and RDTB in the U-Net architecture respectively, `-O' and `-NO' to denote with and without deformation (Offset vs. Non-Offset), and `-$x$' to denote the patch size. For example, `KKDDKK-O-1' denotes the model variant with a patch size of 1, containing 2 RSDTBs (`KK') and 1 RDTB (`D') in the encoder path, and 1 RDTB (`D') and 2 RSDTBs (`KK') in the decoder path that combines the dense and sparse attention with deformable attention (`-O'). `KKKKKK-NO-2' denotes the model variant with a patch size of 2, containing 6 RSDTBs (`KKKKKK') without deformable attention (`-NO'), where only sparse attention is applied.

\begin{figure}[htpb]
    \centering
    \includegraphics[width=4.5in]{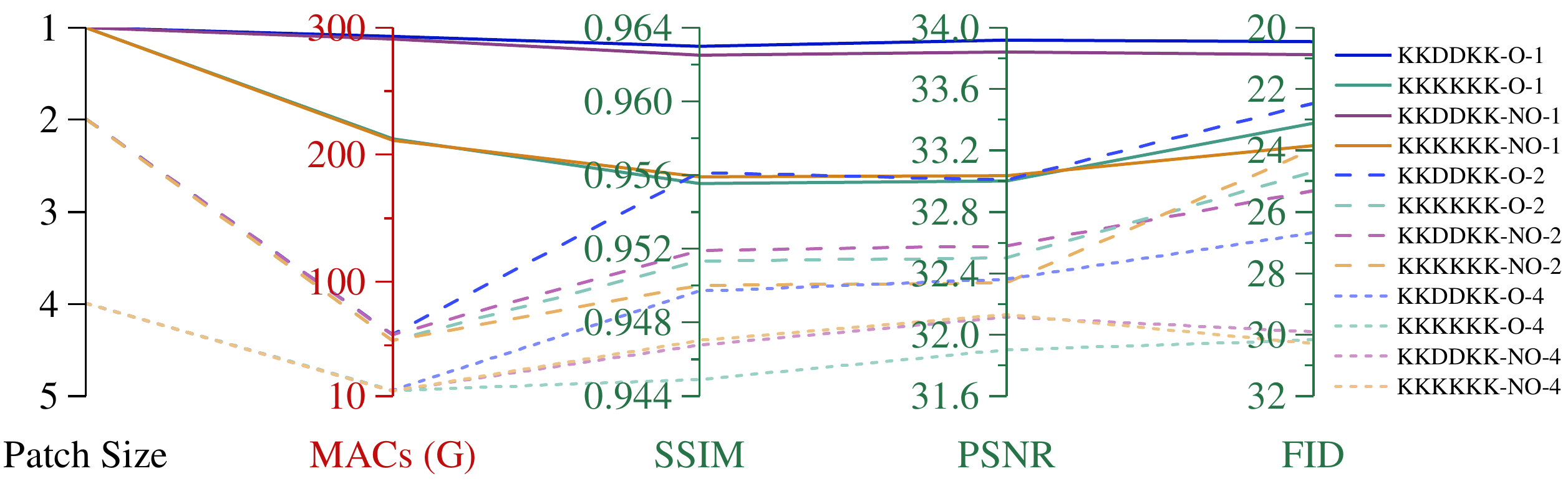}
    \caption{
    Results of our ablation studies. `K' and `D' denote the RSDTB and RDTB in the U-Net architecture, respectively. `-O' and `-NO' represent with and without deformation (Offset vs. Non-Offset). It is of note that our best performed model variant `KKDDKK-O-1' achieved better \textcolor{OliveGreen}{SSIM}, \textcolor{OliveGreen}{PSNR} and \textcolor{OliveGreen}{FID}, but with relatively higher \textcolor{BrickRed}{MACs}.
    }
    \label{fig:FIG_ABLATION}
\end{figure}

Fig.~\ref{fig:FIG_ABLATION} shows that a smaller patch size (with a larger patch resolution) yields better performance (SSIM, PSNR and FID), but entails a larger computational cost (MACs). The utilisation of deformable attention (`-O') improves the reconstruction quality with small computation complexity growth (MACs). For the comparison of block settings in the network, compared to the model only using sparse attention (`K'), the model using both dense (`D') and sparse attention (`K') achieves better performance (SSIM and PSNR), though at the sacrifice of slightly larger computational cost (MACs), especially with lower patch size. The quantitative results table of ablation studies is in the Supplementary.

\section{Discussion and Conclusion}

In this work, we have proposed a novel Swin Deformable Attention U-Net Transformer, i.e., SDAUT, for explainable fast MRI. Our experimental studies have proven our hypotheses:

(1) Deformable attention can strengthen the MRI reconstruction with only limited growth of computational cost. The MACs growth of using deformable attention has been no more than $1\%$.

(2) The incorporation of U-Net architecture can reduce the number of network parameters dramatically and enhance the model performance. Compared to SwinMR, which applies the Swin Transformer backbone with the same number of blocks and Transformer layers, our SDAUT$_{2}$ has achieved comparable performances with only $6.63\%$ computational complexity and our SDAUT$_{1}$ has achieved better performance using $36.59\%$ computational complexity (Table.~\ref{tab:result}).

The combination of dense and sparse attention has gained superior model performance. This is because models with only sparse attention could encounter limitations of global dependency, and models applying only dense attention could increase the computational cost, yielding training difficulties. Fig.~\ref{fig:FIG_ABLATION} has shown that models using both dense and sparse attention (`KKDDKK') outperform models using only sparse attention (`KKKKKK'). However, the `KKDDKK' model setting is more suitable when the patch resolution is not large (e.g., patch size 2 or 4). For smaller patch size of 1, the MACs have been increased by $37.98\%$ (212.37G MACs for `KKKKKK-O-1' vs. 293.02G MACs for `KKDDKK-O-1'), which have lifted a much heavier computing burden.

\begin{figure}[thpb]
    \centering
    \includegraphics[width=4.5in]{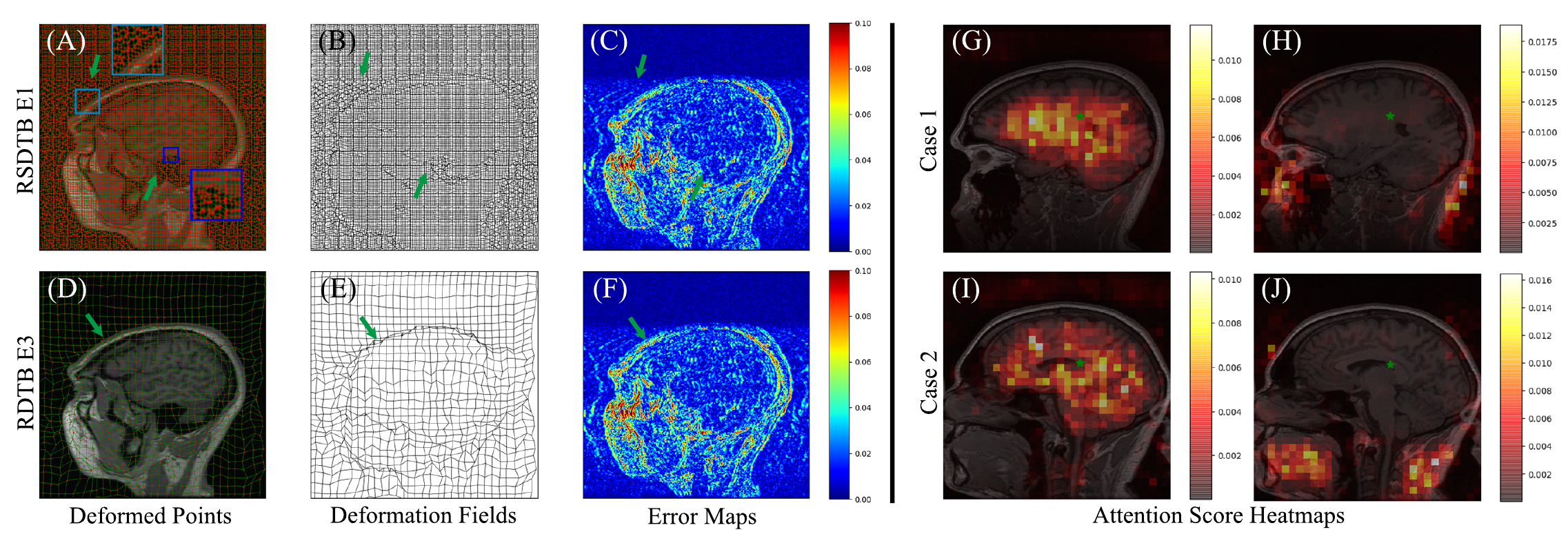}
    \caption{
    (A)(D) Deformed points in RSDTB E1 (Encoder $1^\text{st}$ block) and RDTB E3;
    (B)(E) Deformation fields in RSDTB E1 and RDTB E3;
    (C)(F) Error maps between the undersampled zero-filled MR images and the corresponding ground truth MR images;
    (G)-(J) Attention score heatmaps for a given query (green star) in RDTB E3. 
    }
    \label{fig:FIG_POSITION}
\end{figure}

(3) Our SDAUT has been competent to achieve fast MRI while unveiling model explainability (see Fig. \ref{fig:FIG_SDTL}). Fig.~\ref{fig:FIG_POSITION} (A)-(F) shows the deformed points, deformation fields and corresponding error maps between undersampled zero-filled MR images $x_u$ and corresponding ground truth $x$ (more examples in the Supplementary). We can observe that the deformation field has shown clear differences in information-rich areas (e.g., brain tissue boundaries) and areas with more artefacts (e.g., remnant aliasing on both left and right sides of the head), deciphering higher uncertainties of the reconstruction. Deformable points have been trained to \emph{recognise} and move to the border of the brain, reducing extraneous attention, and therefore improving the model performance.

Fig.~\ref{fig:FIG_POSITION} (G)-(J) shows the attention score heatmaps for a given query (green star) from different heads in RDTB E3 (more examples in the Supplementary). We can observe that different heads have \emph{paid their attention} to different structures (features) in the MRI data, e.g., Fig.~\ref{fig:FIG_POSITION} (G)(I) for the brain tissues and Fig.~\ref{fig:FIG_POSITION} (H)(J) for the vertical high-contrast area (front mouth and back neck regions).
Results have also shown that attention score maps show that a fixed head always focuses on specific structures in MSA, revealing how the multi-head mechanism works. 
Transformer can comprehend different features from the input, and perform attention operations for multiple features using its multi-heads mechanism, which has similar behaviours compared to kernels for different channels in the CNN based models.

\section*{Acknowledgement}

This study was supported in part by the ERC IMI (101005122), the H2020 (952172), the MRC (MC/PC/21013), the Royal Society (IEC\textbackslash NSFC\textbackslash211235), the NVIDIA Academic Hardware Grant Program, the SABER project supported by Boehringer Ingelheim Ltd, and the UKRI Future Leaders Fellowship (MR/V023799/1).

%\bibliographystyle{unsrt}
\bibliographystyle{splncs04}
\bibliography{paper1401}

\end{document}